\documentclass[conference]{IEEEtran}
\IEEEoverridecommandlockouts
\usepackage{amsmath}
\usepackage{amsfonts}
\usepackage{bbding}
\usepackage{amssymb}
\usepackage{array}
\usepackage{subfigure}

\usepackage{graphicx}
\usepackage{subfigure}
\usepackage[named]{algo}
\usepackage{algorithmic}
\usepackage{algorithm}

\usepackage{psfrag}
\usepackage{xfrac}
\usepackage{stfloats}
\usepackage[compress]{cite}
\makeatletter
\renewcommand{\citepunct}{,\penalty\@m\hskip.13emplus.1emminus.1em}
\renewcommand{\citedash}{\hbox{--}\penalty\@m}
\makeatother
\usepackage{setspace}
\usepackage{color}
\allowdisplaybreaks

\usepackage{amsthm}
\usepackage{stfloats}

\newtheorem{rem}{Remark}

\usepackage{hyperref}
\usepackage{bm}
\usepackage{geometry}
\renewcommand{\baselinestretch}{0.93} \normalsize

\begin{document}
\title{\huge Learning to Optimize with Unsupervised Learning: Training Deep Neural Networks for URLLC}

\author{
\IEEEauthorblockN{{Chengjian Sun and Chenyang Yang}} \vspace{0.0cm}
\IEEEauthorblockA{School of Electronics and Information Engineering, Beihang University, Beijing, China\\
Email: \{sunchengjian,cyyang\}@buaa.edu.cn}\vspace{-0.7cm}
}
\maketitle
\begin{abstract}
Learning the optimized solution as a function of environmental parameters is effective in solving numerical optimization in real time for time-sensitive applications. Existing works of learning to optimize train deep neural networks (DNN) with labels, and the learnt solution are inaccurate, which cannot be employed to ensure the stringent quality of service. In this paper, we propose a framework to learn the latent function with unsupervised deep learning, where the property that the optimal solution should satisfy is used as the ``supervision signal'' implicitly. The framework is applicable to both functional and variable optimization problems with constraints. We take a variable optimization problem in ultra-reliable and low-latency communications as an example, which demonstrates that the ultra-high reliability can be supported by the DNN without supervision labels.
\end{abstract}

\section{Introduction}
Many resource allocation problems in wireless systems can be formulated as optimization problems with constraints. The constraints are imposed by the available resources in the system, say maximal bandwidth, or the quality-of-service (QoS) of each user. Depending on the applications, the objective function, constraints and the variables of concern in the formulated problems may not be in the same timescale. If they are in the same timescale, say optimizing power allocation under rate constraint towards a utility that are all instantaneous, then the problem is variable optimization (e.g., \cite{Sun2018TSP}). If they are in different timescales, say in cross-layer optimization \cite{She2018CrossLayer}, then the problem is functional optimization \cite{Zeidler2013Functional}.

Functional optimization problems usually do not have closed-form solutions, and have to be solved numerically, say by the finite element method \cite{Zienkiewicz1977FEM}, which suffers from the curse of dimensionality. Variable optimization problems do not have closed-form solutions if they are non-convex or NP-hard. Whenever the environmental parameters (e.g., channels) change, the numerical solutions have to be re-found, which complexity is unacceptable for real-time processing. To circumvent this issue, a novel approach was proposed in \cite{Sun2018TSP}, which approximates the optimized solution as a function of instantaneous channel gains with neural networks by leveraging the ability of universal approximation. Based on the observation that significant computing efforts are wasted due to repeatedly solving a problem under similar conditions, a deep learning framework was proposed in \cite{Liu2018Lrn2Opt} to find the latent relationship between
flow information and link usage by learning from past computation experience. To learn the optimal predictive resource allocation under the QoS constraint of video-on-demand service, a deep neural network (DNN) was designed in \cite{GY2018}.
By training the DNNs offline, an approximated solution can be obtained with low complexity online \cite{Sun2018TSP, GY2018, Liu2018Lrn2Opt}, say about 1\% of the original numerical optimization \cite{GY2018}. Such an idea of ``learning to optimize" can be regarded as a kind of computing offloading over time, which shifts the computations from online to offline. Since supervised learning is employed in \cite{Sun2018TSP, GY2018, Liu2018Lrn2Opt} where the labels for training are generated from the optimized solutions via simulations, these methods actually make the optimization twice during training: generating labels by repeating the optimization and training the DNN.

The performance of supervised learning highly depends on data, especially labels.
In \cite{Sun2018TSP, Liu2018Lrn2Opt}, the approximated solution can achieve around 90\% performance of the numerical solution. In \cite{GY2018}, even with sophisticated DNN trained with 1,5000 labels generated from the global optimal solution, the QoS of each user can still not be satisfied.
For the application scenarios in \cite{Sun2018TSP, GY2018, Liu2018Lrn2Opt} where the users can be served with best effort, such performance degradation is acceptable. For the applications with stringent QoS constraints such as ultra-reliable and low-latency communications (URLLC), is the DNN-based solutions still useful?

URLLC can support new application scenarios in cellular networks \cite{3GPP2016Scenarios}, which calls for
stringent QoS requirements on the end-to-end (E2E) reliability (e.g., $10^{-5}$ packet loss probability) and latency (e.g., $1$~ms)
 \cite{Adnan2017Realizing}.
The latency in data transmission, channel training, queueing, and decoding have been considered in the literature of URLLC. However, the computing time for numerical optimization is never taken into account, which is not ignorable in the E2E latency.

In this paper, we propose to find the relation between the solution of constrained-optimization and environmental parameters with unsupervised deep learning, where the Karush-Kuhn-Tucker (KKT) conditions serve as the ``supervision" implicitly. By using the property that the optimal solution should satisfy instead of the labels that satisfy the property,
the DNN can learn the relation accurately. By converting variable optimization problems into equivalent functional optimization problems, the proposed framework can be used for both functional and variable optimization.
We take a bandwidth minimization problem subject to the QoS requirement of URLLC as example, aimed to demonstrate how a variable optimization problem can be learnt by the unsupervised learning. Simulation results show that high reliability can be supported by the DNN trained without labels for supervision.


%
%

\section{Learning to Optimize under Constraints without Supervision}
In this section we introduce a framework to learn the solution of optimization problems with constraints without using labels.
We first show how to learn the solution of variable optimization problems as function of parameters. Then, we provide a unified approach for both types of problems.


Consider a general constrained-optimization problem for variables $\bm{x}$ as follows,
\begin{align}   \label{prob:VarOpt}
    \mathop \mathrm{min} \limits_{\bm{x}} \quad& f\left(\bm{x};\bm{\theta}\right) \\
    \text{s.t.} \quad & \bm{C}\left(\bm{x};\bm{\theta}\right) \preceq \bm{0}, \label{con} \tag{\theequation a}
\end{align}
where $f\left(\bm{x};\bm{\theta}\right)$ and $\bm{C}\left(\bm{x};\bm{\theta}\right)$ are the functions of $\bm{x}$ and $\bm{\theta} \!\in\! \mathcal{A}$, $\bm{\theta}$ are environmental parameters such as channel gains, and $\bm{y} \!\preceq\! \bm{0}$ means that each element of vector $\bm{y}$ is less or equal to $0$.


When the optimal solution of the problem in \eqref{prob:VarOpt} does not have closed-form expressions with respect to (w. r. t.) to $\bm{\theta}$, numerical solution has to be found again and again whenever one or more elements in vector $\bm{\theta}$ changes. This is not affordable for time-sensitive applications. To reduce the online computation time, an emerging technique is to find the relation between the optimal solution and parameters, i.e., $\bm{x}^*(\bm{\theta})$, by offline training with supervised learning \cite{Sun2018TSP, GY2018, Liu2018Lrn2Opt}. In the sequel, we introduce a framework to find $\bm{x}^*(\bm{\theta})$ with unsupervised learning.


To embody the relation between $\bm{x}$ and $\bm{\theta}$, we convert the variable optimization problem in \eqref{prob:VarOpt} into the following functional optimization problem:
\begin{align}   \label{prob:VarOptb}
    \mathop \mathrm{min} \limits_{\bm{x}(\bm{\theta})} \quad& \mathcal{F}\left[f\left(\bm{x}(\bm{\theta});\bm{\theta}\right)\right] \\
    \text{s.t.} \quad & \eqref{con}, \  \forall {\bm{\theta} \in \mathcal{A}}, \nonumber
\end{align}
where $\mathcal{F}[\cdot]$ is a functional with positive variation. In Appendix \ref{App:Upgrade}, we prove that the problems in \eqref{prob:VarOptb} and  \eqref{prob:VarOpt} are equivalent in the sense that they yield the same optimal solution.

As an illustration, we consider $\mathcal{F}[\cdot]$ as the mathematical expectation, i.e., we take the expectation of the objective function in the problem in \eqref{prob:VarOpt} over $\bm{\theta}$, which is
\begin{align}   \label{prob:FuncOpt}
    \mathop \mathrm{min} \limits_{\bm{x}(\bm{\theta})} \quad& \int_{\bm{\theta} \in \mathcal{A}} {f\left(\bm{x}(\bm{\theta});\bm{\theta}\right) p(\bm{\theta}) \mathrm{d} \bm{\theta}} \\
    \text{s.t.} \quad & \eqref{con}, \  \forall {\bm{\theta} \in \mathcal{A}}, \nonumber
\end{align}
where $p(\bm{\theta}) \!>\! 0$ is the probability density function (PDF) of parameters $\bm{\theta}$, which can be arbitrary function that ensures the objective function to be integrable.


When the Slater's condition holds, the problem in \eqref{prob:FuncOpt} is equivalent to the following problem \cite{boyd},
\begin{small}
\begin{align}   \label{prob:FuncOptLag}
    \mathop \mathrm{min} \limits_{\bm{x}(\bm{\theta})} \mathop \mathrm{max} \limits_{\bm{\lambda}(\bm{\theta})} \ & L \!\triangleq\! \int_{\bm{\theta} \in \mathcal{A}} \!\!\!\!\!\!{\left[f\left(\bm{x}(\bm{\theta});\bm{\theta}\right) \!+\! \bm{\lambda}^\mathrm{T}(\bm{\theta}) \bm{C}\left(\bm{x}(\bm{\theta});\bm{\theta}\right)\right] p(\bm{\theta}) \mathrm{d} \bm{\theta}} \\
    \text{s.t.} \  & \bm{\lambda}(\bm{\theta}) \succeq \bm{0}, \  \forall {\bm{\theta} \in \mathcal{A}}, \label{con:lambda} \tag{\theequation a}
\end{align}
\end{small}
where $L$ is the Lagrange function, $\bm{\lambda}(\bm{\theta})$ is the vector of Lagrange multipliers, ${(\cdot)}^\mathrm{T}$ denotes transpose, and $\bm{y} \!\succeq\! \bm{0}$ means that each element of vector $\bm{y}$ is greater or equal to $0$.

The optimal solution of the problem in \eqref{prob:FuncOptLag} should satisfy its KKT conditions, which can be derived as follows,
\begin{align}
    & \frac{\mathrm{\delta} L}{\mathrm{\delta} \bm{x}(\bm{\theta})} = \left(\nabla_{\bm{x}} f \!+\! \nabla_{\bm{x}} \bm{C} \bm{\lambda}\right) p(\bm{\theta}) = \bm{0}, \  \forall {\bm{\theta} \in \mathcal{A}}, \label{KKT_x} \\
    & \bm{\lambda} \star \bm{C} = \bm{0}, \  \forall {\bm{\theta} \in \mathcal{A}}, \label{KKT_lambda} \\
    & \eqref{con}, \ \eqref{con:lambda}, \  \forall {\bm{\theta} \in \mathcal{A}}, \nonumber
\end{align}
where $\frac{\mathrm{\delta} L}{\mathrm{\delta} \bm{x}(\bm{\theta})}$ is the variation of $L$ w. r. t. function $\bm{x}(\bm{\theta})$, and $\star$ denotes element wise multiplication.

The optimal solution of the problem in \eqref{prob:FuncOpt} is hard to find from the KKT conditions, because it is the
functions $\bm{x}(\bm{\theta})$ and $\bm{\lambda}(\bm{\theta})$ that need to be optimized, which can be interpreted as the vectors with infinite elements.

Thanks to the universal approximation theorem \cite{Hornik1989UnivApprox}, we can approximate $\bm{x}(\bm{\theta})$ and $\bm{\lambda}(\bm{\theta})$ with neural networks $\mathcal{N}_x (\bm{\theta};\bm{\omega}_x)$ and $\mathcal{N}_\lambda (\bm{\theta};\bm{\omega}_\lambda)$, respectively, where $\bm{\omega}_x$ and $\bm{\omega}_\lambda$ are finite network parameters. Then, by replacing $\bm{x}(\bm{\theta})$ and $\bm{\lambda}(\bm{\theta})$ with $\hat{\bm{x}} \!\triangleq\! \mathcal{N}_x (\bm{\theta};\bm{\omega}_x)$ and $\hat{\bm{\lambda}} \!\triangleq\! \mathcal{N}_\lambda (\bm{\theta};\bm{\omega}_\lambda)$, the problem in \eqref{prob:FuncOptLag} can be re-written as,
\begin{align}   \label{prob:FuncOptLagParam}
    \mathop \mathrm{min} \limits_{\bm{\omega}_x} \mathop \mathrm{max} \limits_{\bm{\omega}_\lambda} \quad & \hat{L} \triangleq \int_{\bm{\theta} \in \mathcal{A}} {\left(\hat{f} + \hat{\bm{\lambda}}^\mathrm{T} \hat{\bm{C}}\right) p(\bm{\theta}) \mathrm{d} \bm{\theta}} \\
    \text{s.t.} \quad & \hat{\bm{\lambda}} \succeq \bm{0}, \  \forall {\bm{\theta} \in \mathcal{A}}, \nonumber
\end{align}
where $\hat{f} \!\triangleq\! f\left(\hat{\bm{x}};\bm{\theta}\right)$ and $\hat{\bm{C}} \!\triangleq\! \bm{C}\left(\hat{\bm{x}};\bm{\theta}\right)$.
The KKT conditions of the problem in \eqref{prob:FuncOptLagParam} can be derived as,
\begin{align}
    & \nabla_{\bm{\omega}_x} \hat{L} \!=\! \int_{\bm{\theta} \in \mathcal{A}} {\nabla_{\bm{\omega}_x} \hat{\bm{x}} \left(\nabla_{\bm{x}} \hat{f}\!+\! \nabla_{\bm{x}} \hat{\bm{C}} \hat{\bm{\lambda}}\right) p(\bm{\theta}) \mathrm{d} \bm{\theta}} \!=\! \bm{0}, \label{KKTParam_x} \\
    &\nabla_{\bm{\omega}_\lambda} \hat{L} \!=\! \int_{\bm{\theta} \in \mathcal{A}} {\nabla_{\bm{\omega}_\lambda} \hat{\bm{\lambda}} \hat{\bm{C}} p(\bm{\theta}) \mathrm{d} \bm{\theta}} \!=\! \bm{0}, \label{KKTParam_lambda} \\
    & \hat{\bm{C}} \preceq \bm{0}, \ \hat{\bm{\lambda}} \succeq \bm{0}, \ \forall {\bm{\theta} \in \mathcal{A}}, \nonumber
\end{align}
where the gradients $\nabla_{\bm{\omega}_x} \hat{\bm{x}}$ and $\nabla_{\bm{\omega}_\lambda} \hat{\bm{\lambda}}$ can be computed via back propagation.

\begin{rem}
\em{To learn the optimal neural networks, we can take $\hat{L}$ as the loss function, and train $\bm{\omega}_x$ and $\bm{\omega}_\lambda$ with gradient based methods such as stochastic gradient descent (SGD) method. Since no label is used for supervision, the proposed learning method is unsupervised, where the KKT conditions implicitly serve as the ``supervised signal".}
\end{rem}
\begin{rem}
    \em{Comparing the KKT conditions of the problem in \eqref{prob:FuncOptLag} and the problem in \eqref{prob:FuncOptLagParam}, we can find that the conditions in \eqref{KKT_x} and \eqref{KKT_lambda} are tighter than the conditions in \eqref{KKTParam_x} and \eqref{KKTParam_lambda}. This is because the former ones should be satisfied for each value of $\bm{\theta}$, while the later ones only need to be satisfied on average over $\bm{\theta}$. Because the numbers of equations in \eqref{KKTParam_x} and \eqref{KKTParam_lambda} are respectively equal to the numbers of elements in $\bm{\omega}_x$ and $\bm{\omega}_\lambda$, increasing the scale of $\mathcal{N}_x (\bm{\theta};\bm{\omega}_x)$ and $\mathcal{N}_\lambda (\bm{\theta};\bm{\omega}_\lambda)$ can make the conditions in \eqref{KKTParam_x} and \eqref{KKTParam_lambda} tighter and closer to the original KKT conditions in \eqref{KKT_x} and \eqref{KKT_lambda}. When the dimension of $\bm{\omega}_x$ and $\bm{\omega}_\lambda$ is sufficient large, the two functions $\bm{x}(\bm{\theta})$ and $\bm{\lambda}(\bm{\theta})$ can be approximated by DNN
    accurately.}
\end{rem}
\begin{rem}
\em{For the constrained-optimization problem for functions, we can still use the unsupervised learning framework to find the relation between the optimal solution and the parameters. The only difference lies in the inputs of the neural networks, which not only consist of the parameters but also the input variables of the functions to be optimized.}
\end{rem}

\section{An Example Problem of URLLC}
In this section, we illustrate how to learn the constrained variable optimization problem with a QoS-constrained bandwidth minimization problem for URLLC.
\subsection{System Models} \label{sec:system_model}
Consider an uplink (UL) orthogonal frequency division multiple access system, where a base station (BS) with $N_\mathrm{t}$ antennas serves $K$ single-antenna users. The maximal transmit power of each user is $P_\mathrm{max}$. The bandwidth allocated to the $k$th user is $W_k$.  Since the packet size $u$ in URLLC is typically small (e.g., $20$~bytes \cite{3GPP2016Scenarios}), the bandwidth required for transmitting each packet is less than the channel coherence bandwidth. Therefore, the channel is flat fading.

Time is discretized into frames, each with duration $T_\mathrm{f}$. In each frame, multiple packets are generated at the $k$th user and will be sent to the BS. The duration for UL data transmission in one frame is $\tau$ and the duration for channel training is $T_\mathrm{f} \!-\! \tau$. Since the E2E delay requirement in URLLC is typically shorter than the channel coherence time, the channel is quasi-static and time diversity cannot be exploited. To guarantee the transmission reliability within the delay bound, we consider frequency hopping, where each user is assigned with different subchannels in adjacent frames. When the frequency interval between adjacent subchannels is larger than the coherence bandwidth, the small scale channel gains of a user among frames are mutually independent.

\subsubsection{Achievable Rate in Finite Blocklength Regime}
In URLLC, the blocklength of channel coding is short due to the short transmission duration, and hence the impact of decoding errors on reliability cannot be ignored. Shannon's capacity formula cannot be employed to characterize the probability of decoding errors \cite{Gross2015Delay}. The achievable rate in finite blocklength regime is required. In quasi-static flat fading channels, when the instantaneous
channel gain is available at the transmitter and receiver, the achievable rate of the $k$th user (in packets/frame) can be accurately approximated by \cite{Yury2014Quasi},
\begin{align}   \label{eq:Srv}
    s_k \!\approx\! \frac{\tau W_k}{u \ln{2}} \left[\ln\left(1 \!+\! \frac{\alpha_k g_k P_\mathrm{max}}{ N_0 W_k}\right) \!-\! \sqrt{\frac{V_k}{\tau W_k}} Q_\mathrm{G}^{-1}\!\left({\varepsilon^\mathrm{c}_k}\right)\right],
\end{align}
where $\varepsilon^\mathrm{c}_k$ is the decoding error probability of the $k$th user, $\alpha_k$ and $g_k$ are the large-scale channel gain and small-scale channel gain of the $k$th user, respectively, $N_0$ is the single-side noise spectral density, $Q_\mathrm{G}^{-1} (x)$ is the inverse of the Gaussian Q-function, and $V_k=1 \!-\! 1/{(1+\frac{\alpha_k g_k P_\mathrm{max}}{N_0 W_k })}^2$ \cite{Yury2014Quasi}.

Although the achievable rate is in closed-form, it is still too complicated to obtain graceful results. As shown in \cite{Gross2015Delay}, if the signal-to-noise ratio (SNR) $\frac{\alpha_k g_k P_\mathrm{max}}{N_0 W_k } \ge$ $5$~dB, $V_k \!\approx\! 1$ is accurate. Since high SNR is required to ensure ultra-high reliability and ultra-low latency, such approximation is reasonable. Even when the SNR is not high, we can obtain a lower bound of the achievable rate by substituting $V_k \!\approx\! 1$ into $s_k$. Then, when the required $\varepsilon^\mathrm{c}$ is satisfied with the lower bound, it can also be satisfied with the achievable rate in \eqref{eq:Srv}.

\subsubsection{Quality-of-Service}
When the service rate is random due to the channel fading, the packets may not be served immediately and will accumulate into a queue in the buffer at the user. The QoS requirements of URLLC can be characterized by the delay bound $D_\mathrm{max}$ and the overall packet loss probability $\varepsilon_\mathrm{max}$. The downlink transmission delay is subtracted from the E2E delay in this paper, since it has been studied in \cite{She2018CrossLayer}. Thus, herein $D_\mathrm{max}$ is the UL delay, which consists of the queueing delay (denoted as $D^\mathrm{q}_k$ for the $k$th user), transmission delay $D^\mathrm{t}$ and decoding delay $D^\mathrm{c}$. All these delay components are measured in frames. $D^\mathrm{t}$ and $D^\mathrm{c}$ are constant values \!\cite{Condoluci2017Reserv}. Due to the random packet arrival, $D^\mathrm{q}_k$ is random. To ensure the delay requirement, $D^\mathrm{q}_k$ should be bounded by $D^\mathrm{q}_\mathrm{max} \!\triangleq\! D_\mathrm{max} \!-\! D^\mathrm{t} \!-\! D^\mathrm{c}$. If the queueing delay of a packet exceeds $D^q_\mathrm{max}$, the packet will be useless.


Denote $\varepsilon^\mathrm{q}_k \!\triangleq\! \Pr\{D^\mathrm{q}_k \!>\! D^\mathrm{q}_\mathrm{max}\}$ as the queueing delay violation probability.
Then, the overall reliability requirement can be characterized by
\begin{align}   \label{eq:Relia}
    1-(1-\varepsilon^\mathrm{c}_k)(1-\varepsilon^\mathrm{q}_k) \approx \varepsilon^\mathrm{c}_k+\varepsilon^\mathrm{q}_k \leq \varepsilon_\mathrm{max}.
\end{align}
This approximation is very accurate, because the values of $\varepsilon^\mathrm{c}_k$ and $\varepsilon^\mathrm{q}_k$ are very small in URLLC.

\subsection{Minimizing the Bandwidth to Ensure the QoS}
 In what follows, we show how to find the minimal bandwidth required by each user to support the QoS requirement in URLLC as a function of the large-scale channel gain. We first show how to find the optimal solution for each given value of the large-scale channel gain, and then show how to learn the optimal solutions for arbitrary values of the large-scale channel gains in an unsupervised manner.

 \subsubsection{Problem Formulation}
 Since the achievable rate in \eqref{eq:Srv} depends on the small-scale channel gain, the packet service rate of each user is random. In this case, effective capacity can be applied to analyze the queueing delay \cite{Lingjia2007TIT},\footnote{We have validated that effective capacity can be applied in URLLC, but do not show the results due to the space limitation.} with which an upper bound of the queueing delay violation probability of the $k$th user can be expressed as
 \begin{align}   \label{eq:QUpp}
    \varepsilon^\mathrm{q}_k < e^{-\theta_k C^\mathrm{E}_k D^\mathrm{q}_\mathrm{max}},
\end{align}
where $C^\mathrm{E}_k$ is the effective capacity of the service process of the $k$th user, and  $\theta_k$ is the QoS exponent. Since the small-scale channel gains of a user are independent among frames owing to frequency hopping, the effective capacity of the $k$th user can be expressed as \cite{Tang2007TWC}
\begin{align}   \label{eq:EC}
    C^\mathrm{E}_k = - \frac{1}{\theta_k} \ln{\mathbb{E}_{g_k} \left\{e^{- \theta_k s_k}\right\}} \; \text{(packets/frame)},
\end{align}
where the expectation is taken over small-scale channel gains.

If the upper bound in \eqref{eq:QUpp} is satisfied, then the queueing delay requirement ($D^\mathrm{q}_\mathrm{max}$, $\varepsilon^\mathrm{q}_k$) can be satisfied. Furthermore, both the delay requirement and the overall reliability requirement in \eqref{eq:Relia} (i.e., the QoS requirement) can be satisfied if
\begin{align}   \label{eq:ReliaCnsrv}
    \varepsilon^\mathrm{c}_k + e^{-\theta_k C^\mathrm{E}_k D^\mathrm{q}_\mathrm{max}} = \varepsilon_\mathrm{max}.
\end{align}
As shown in \cite{She2018CrossLayer}, the optimal values of the packet loss probabilities are in the same order of magnitude. Therefore, we set $\varepsilon^\mathrm{c}_k \!=\! e^{-\theta_k C^\mathrm{E}_k D^\mathrm{q}_\mathrm{max}} \!=\! \varepsilon_\mathrm{max}/2$ for simplicity.

To satisfy the upper bound in \eqref{eq:QUpp}, the effective capacity should not be lower than the packet generation rate, i.e., $C^\mathrm{E}_k \!\geq\! m_k$, where $m_k$ is the number of packets generated at the $k$th user in each frame.
Noticing that if the upper bound is loose, then more bandwidth is required to ensure the QoS conservatively. Hence, we optimize the QoS exponent together with the bandwidth allocated to each user to minimize the total bandwidth, which yields a tighter upper bound.
For given large scale channel gains, the optimal bandwidth allocation problem that minimizes the total bandwidth required to ensure the QoS of every user can be formulated as,
\begin{small}
\begin{align}   \label{prob:RA}
    \mathop \mathrm{min} \limits_{W_k, \theta_k} \quad& \sum_{k=1}^{K} {W_k} \\
    \text{s.t.} \quad
    & - \frac{1}{\theta_k} \ln{\mathbb{E}_{g_k} \left\{e^{- \theta_k s_k}\right\}} \geq m_k, \label{con:Q} \tag{\theequation a} \\
    & s_k \!=\! \frac{\tau W_k}{u \ln{2}} \left[\ln\!\left(1\!+\!\frac{\alpha_k g_k P_\mathrm{max}}{ N_0 W_k}\right) \!-\!\frac{Q_\mathrm{G}^{-1}\!\left({\varepsilon_\mathrm{max}/2}\right)}{\sqrt{\tau W_k}}\right], \label{con:Srv} \tag{\theequation b} \\
    & e^{-\theta_k C^\mathrm{E}_k D^\mathrm{q}_\mathrm{max}} = \varepsilon_\mathrm{max}/2, \label{con:ReliaQ} \tag{\theequation c} \\
    & W_k > 0, \nonumber
\end{align}
\end{small}
where \eqref{con:Q} is the queueing delay requirement, \eqref{con:Srv} is the achievable packet rate in \eqref{eq:Srv} to support the decoding reliability requirement  $\varepsilon^\mathrm{c}_k$, and \eqref{con:ReliaQ} is from \eqref{eq:ReliaCnsrv} to ensure the overall reliability requirement.

Since constraints \eqref{con:Q}, \eqref{con:Srv} and \eqref{con:ReliaQ} of a user do not rely on the bandwidth allocated to and the QoS exponents of other users, problem \eqref{prob:RA} can be equivalently decomposed into multiple problems each finding the minimum bandwidth required by each user. In what follows, we only consider the single user scenario of problem \eqref{prob:RA} and omit the subscript $k$ for notational simplicity. Then, the relation needs to learn is $W^*(\alpha)$, i.e., the parameter in \eqref{prob:VarOpt} is $\alpha$.

\subsubsection{Finding the Optimal Solution Given $\alpha$}
To provide a baseline for the unsupervised learning method,
 in what follows we show how to find the optimal solution with given values of the large-scale channel gain.
Since less bandwidth is required if the queueing delay requirement is looser, the optimal solution of the problem in \eqref{prob:RA} should be obtained when the equality in \eqref{con:Q} holds. Then, from the equalities in \eqref{con:Q} and \eqref{con:ReliaQ}, the optimal QoS exponent can be solved as
\begin{align}   \label{eq:theta}
   \theta^* \!= -\frac{\ln{(\varepsilon_\mathrm{max}/2)}}{m D^\mathrm{q}_\mathrm{max}},
\end{align}
and the optimal bandwidth can be found by finding the solution of the equalities in \eqref{con:Q} and \eqref{con:Srv}.

If effective capacity can be derived as an close-form expression, say for large scale antenna systems \cite{ECTCOM15}, then numerical solution can be found for the problem in \eqref{prob:RA}. For general wireless systems where the effective capacity is not with the closed form, the constraint in \eqref{con:Q} does not have closed-form expression. For such kind of problems, we can resort to stochastic optimization methods such as SGD for minimization problems and stochastic gradient ascent (SGA)
for maximization problems. To obtain an unbiased gradient estimation with stochastic gradient methods, the expectations in the objective function and constraints of a problem should not be in nonlinear forms. Thus, we transform \eqref{con:Q}  into an equivalent form that is linear to the expectation.
Further substituting $\theta^*$, the QoS constraint becomes
\begin{align}   \label{con:QLnr}
    \mathbb{E}_{g_k} \left\{e^{- \theta^* s}\right\} - e^{- \theta^* m} \leq 0.
\end{align}

Due to the expectation in \eqref{con:QLnr} and the complex expression of the achievable rate in \eqref{con:Srv}, the monotonicity of the function in the left hand side of \eqref{con:QLnr} is hard to analyze. In concept, the achievable rate should increase with the bandwidth. However, this may not be true when the small-scale channel gain is very small (lower than $-10$~dB) due to the approximation $V \!\approx\! 1$. Fortunately, since very small values of the small-scale channel gain rarely occur (e.g., $\Pr\{g \!<\! 0.1\} \!<\! 10^{-12}$ when $N_\mathrm{t} \!\geq\! 8$ for Rayleigh fading channels), the impact can be ignored after taking the expectation. Therefore, it is reasonable to assume that the left-hand side of \eqref{con:QLnr} decreases with $W$. Then, the optimal bandwidth allocation can be found with stochastic optimization through the following iterations,
\begin{align}   \label{opt:W}
    W^{(t+1)} = {\left[W^{(t)} + \phi(t) \left(e^{- \theta^* s^{(t)}} - e^{-\theta^* B^\mathrm{E}}\right)\right]}^+,
\end{align}
where ${\left[x\right]}^+ \!\triangleq\! \max\!{\left\{x,0\right\}}$ ensures the results to be positive, $\phi(t) \!>\! 0$ is the step size, and $s^{(t)}$ is the achievable rate computed from the realization of $g$ in the $t$th iteration according to \eqref{con:Srv}. With the aforementioned assumption (which is true as we validated via simulations) and $\phi(t) \!\sim\! \mathcal{O}\!\left(\frac{1}{t}\right)$, $\{W^{(t)}\}$ converges to the unique optimal bandwidth \cite{Bottou1998Stochastic}.
\subsubsection{Learning $W^*(\alpha)$ without Supervision}
In the sequel we show how to employ unsupervised learning to find the approximated relation between the optimal bandwidth allocated to a user and the large-scale channel gain of the user. The QoS exponent is not learned since it has analytical solution as in \eqref{eq:theta}. We first transform the bandwidth optimization problem to the following functional optimization problem,
\begin{align}   \label{prob:RA_FuncOpt}
    \mathop \mathrm{min} \limits_{W(\alpha)} \quad& \mathbb{E}_{\alpha} \left\{W(\alpha)\right\} \\
    \text{s.t.} \quad & \eqref{con:Srv}, \  \eqref{con:QLnr}, \  W(\alpha) > 0. \nonumber
\end{align}
It is not hard to show that the  Slater¡¯s condition holds, i.e., this problem is equivalent to the following problem,
\begin{small}
\begin{align}
    \mathop \mathrm{min} \limits_{W(\alpha)} \mathop \mathrm{max} \limits_{\lambda(\alpha)} \quad& L \triangleq \mathbb{E}_{\alpha} \left\{W(\alpha) + \lambda(\alpha) \left( \mathbb{E}_{\bm{g}} \!\left\{\!e^{- \theta^* s}\!\right\} \!-\! e^{-\theta^* m} \right)\right\} \nonumber \\
    \text{s.t.} \quad & \eqref{con:Srv}, \  W(\alpha) > 0, \  \lambda(\alpha) > 0, \nonumber
\end{align}
\end{small}
where $\lambda(\alpha)$ is the Lagrange multiplier.

Then, we approximate $W(\alpha)$ and $\lambda(\alpha)$ with fully connected  neural networks $\hat{W} \!\triangleq\! \mathcal{N}_W \left(\alpha;\bm{\omega}_W\right)$ and $\hat{\lambda} \!\triangleq\! \mathcal{N}_\lambda \left(\alpha;\bm{\omega}_\lambda\right)$, respectively. By applying \texttt{Softplus} in the output layers in both neural networks, $\hat{W}$ and $\hat{\lambda}$ are automatically positive. Taking $\hat{L}$ as the loss function, which is obtained by substituting $W(\alpha) \!\approx\! \hat{W}$ and $\lambda(\alpha) \!\approx\! \hat{\lambda}$ into $L$, we can train $\bm{\omega}_W$ and $\bm{\omega}_\lambda$ respectively with SGD and SGA,
\begin{align}
    \bm{\omega}_W^{(t+1)} \!&=\! \bm{\omega}_W^{(t)} \!\!-\! \phi(t) \nabla_{\bm{\omega}_W} \hat{L}^{(t)} \nonumber \\
    &=\! \bm{\omega}_W^{(t)} \!\!-\! \phi(t) \nabla_{\bm{\omega}_W} \mathcal{N}_W\!\left(\alpha^{(t,n)};\bm{\omega}_W^{(t)}\right) \frac{\mathrm{d} \hat{L}^{(t)}}{\mathrm{d} W}, \label{trn:ParaW} \\
    \bm{\omega}_\lambda^{(t+1)} \!&=\! \bm{\omega}_\lambda^{(t)} \!\!+\! \phi(t) \nabla_{\bm{\omega}_\lambda} \hat{L}^{(t)} \nonumber \\
    &=\! \bm{\omega}_\lambda^{(t)} \!\!+\! \phi(t) \nabla_{\bm{\omega}_\lambda} \mathcal{N}_\lambda\!\left(\alpha^{(t,n)};\bm{\omega}_W^{(t)}\right) \frac{\mathrm{d} \hat{L}^{(t)}}{\mathrm{d} \lambda}, \label{trn:ParaLambda}
\end{align}
where $\hat{L}^{(t)} \!\triangleq\! \frac{1}{N_\mathrm{b}} \sum_{n=1}^{N_\mathrm{b}} {\left[\hat{W} \!+\! \hat{\lambda} \!\left( e^{-\! \theta \hat{s}^{(t,n)}} \!\!-\! e^{-\theta m} \right)\right]}$ is the realization of $\hat{L}$ in the $t$th iteration, $N_\mathrm{b}$ is the number of realizations in each batch, $\alpha^{(t,n)}$ and $\hat{s}^{(t,n)}$ are the $n$th realizations of the large-scale channel gain and the achievable rate in the $t$th iteration, respectively.
The gradient matrixes of the neural networks w. r. t. the parameters $\nabla_{\bm{\omega}_W} \mathcal{N}_W\left(\alpha^{(t,n)};\bm{\omega}_W^{(t)}\right)$ and $\nabla_{\bm{\omega}_\lambda} \mathcal{N}_\lambda\left(\alpha^{(t,n)};\bm{\omega}_\lambda^{(t)}\right)$ can be computed by backward propagation, $\frac{\mathrm{d} \hat{L}^{(t)}}{\mathrm{d} W} \!=\! 1 \!-\! \frac{1}{N_\mathrm{b}} \sum_{n=1}^{N_\mathrm{b}} {\hat{\lambda}^{(t)} \theta \frac{\partial \hat{s}^{(t,n)}}{\partial \hat{W}} e^{- \theta \hat{s}^{(t,n)}}}$ and $\frac{\mathrm{d} \hat{L}^{(t)}}{\mathrm{d} \lambda} \!=\! \frac{1}{N_\mathrm{b}} \!\sum_{n=1}^{N_\mathrm{b}}{\!\left( e^{-\! \theta^* \hat{s}^{(t,n)}} \!\!-\! e^{-\theta^* m} \right)}$.
\begin{rem}
\em{From the iteration of the parameters of the neural network for Lagrange multiplier
in \eqref{trn:ParaLambda}, we can find that if the iteration converges then
the QoS constraint in \eqref{con:QLnr} will be satisfied. Extensive simulation results show that the iteration always converges for this example problem, despite that it is hard to prove the convergency.}
\end{rem}

%

\section{Simulation Results}    \label{sec:Results}
In this section, we first show the tradeoff between the minimal bandwidth required to ensure the QoS. Then, we show the performance of the approximated optimal solution obtained with the unsupervised learning.

Without lose of generality, we consider a single user case in a single cell with radius of 250 m and path loss model $10\lg(\alpha) \!=\! 35.3+37.6 \lg(d)$. The simulation setup and fine-tuned hyper-parameters for the neural network are listed in Table \ref{tab:SimParam}, unless otherwise specified. We use \texttt{Softplus} in the output layers in all DNNs, and use \texttt{TanH} in the hidden layers as an example activation function.

\begin{table}[htbp]
	\footnotesize
	\renewcommand{\arraystretch}{1.3}
	\caption{Simulation Parameters and Hyper-parameters}	\label{tab:SimParam}
	\begin{center}\vspace{-0.2cm}
	\begin{tabular}{|p{4.5cm}|p{2.7cm}|}
		\hline
        Overall packet loss probability $\varepsilon_\mathrm{max}$ & $10^{-5}$ \\ \hline
        UL delay bound $D_\mathrm{max}$ & $10$~frames ($1$~ms) \\ \hline
		Duration of each frame $T_\mathrm{f}$ & $0.1$~ms \\ \hline
		Duration of UL transmission  $\tau$ & $0.05$~ms \\ \hline
        Transmission delay $D^\mathrm{t}$ & $1$~frame \cite{Condoluci2017Reserv}\\ \hline
        Decoding delay $D^\mathrm{c}$ & $1$~frame \cite{Condoluci2017Reserv}\\ \hline
        Maximal transmit power of BS $P_\mathrm{max}$ & $23$~dBm \\ \hline
        Number of antennas $N_\mathrm{t}$ & 8 \\ \hline
		Single-sided noise spectral density $N_0$ & $-173$~dBm/Hz \\ \hline
		Packet size $u$ & $20$~bytes ($160$~bits) \cite{3GPP2016Scenarios} \\ \hline
        Packets generated in each frame $m$ & $1$~packet \\ \hline
        Learning rate $\phi(t)$ & $0.1$ \\ \hline
        Number of hidden layers & $4$ \\ \hline
        Number of neurons in each layer & $8$ \\ \hline
        Batch size $N_\mathrm{b}$ & $100$ \\ \hline
	\end{tabular}
	\end{center}
	\vspace{-0.2cm}
\end{table}

\begin{table}[htbp]
	\small
	\renewcommand{\arraystretch}{1.3}
	\caption{Minimal Bandwidth Required by Different Reliability Requirements (MHz), $W^*$.}	\label{tab:Relia}
	\begin{center} \vspace{-0.4cm}
    \begin{tabular}{|l|c|c|c|c|}
    \hline
    Required $\varepsilon_\mathrm{max}$ & $10^{-4}$ & $10^{-5}$ & $10^{-6}$ & $10^{-7}$ \\ \hline
    $d \!=\! 250$~m & $0.512$ & $0.527$ & $0.542$ & $0.556$ \\ \hline
    $d \!=\! 50$~m & $0.204$ & $0.208$ & $0.211$ & $0.213$ \\ \hline
    \end{tabular}
    \end{center}
	\vspace{-0.7cm}
\end{table}

Table \ref{tab:Relia} shows the minimal bandwidth required to ensure the required QoS, where several overall packet loss probability requirements are considered. The results are obtained by finding the optimal solution given $\alpha$ (i.e., the distance $d$). From the results we can find that the minimal bandwidth is insensitive to the required reliability. For the user located at cell-centre and the user at cell-edge, only around $3\%$ of additional bandwidth is required to enhance the reliability by one order of magnitude.

To evaluate the results of $W^*(\alpha)$ learned without supervision, $\mathcal{N}_W \left(\alpha;\bm{\omega}_W\right)$ and $\mathcal{N}_\lambda \left(\alpha;\bm{\omega}_\lambda\right)$ are trained according to the iterations in \eqref{trn:ParaW} and \eqref{trn:ParaLambda}. In each iteration, $N_\mathrm{b}$ realizations are generated, where the small-scale channel gain are randomly generated from Rayleigh distribution, and the large-scale channel gains are computed from the path loss model with user-BS distance $d$ uniformly distributed in $50 \sim 250$~m.

\begin{figure}[htbp]
	\vspace{-0.4cm}
	\centering
	\begin{minipage}[t]{0.46\textwidth}
	\includegraphics[width=1\textwidth]{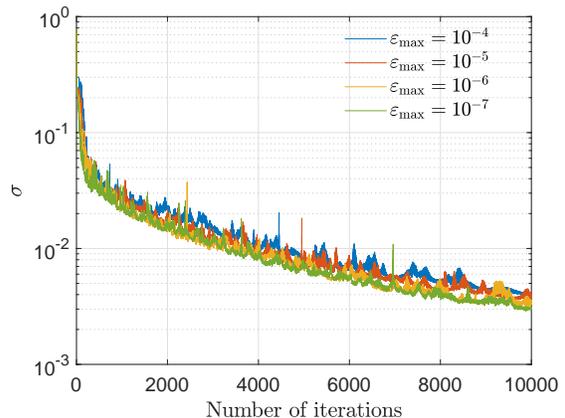}
	\end{minipage}
	\vspace{-0.3cm}
	\caption{Convergence of the proposed unsupervised learning.}\vspace{-0.2cm}
	\label{fig:LrnCrvs}	\vspace{-0.1cm}
	\vspace{-0.1cm}
\end{figure}

Fig. \ref{fig:LrnCrvs} shows how the approximation accuracy, defined as $\sigma \!\triangleq\! \sqrt{{\mathbb{E}_\alpha \big\{{(\hat{W} \!-\! W^*)}^2\big\}} \big/ {\mathbb{E}_\alpha \big\{{(W^*)}^2\big\}}}$, changes with iterations, where $W^*$ is obtained by finding the optimal solution given $\alpha$ with \eqref{opt:W}. Each curve is obtained by averaging over $100$ times of training. From the results we can find that the learnt solution can well approximate the optimal solution, and the convergence speed is insensitive to the reliability requirement. Specifically, $\sigma<1\%$ after around $6\,000$ iterations for a wide range of the reliability requirement.

Due to the stochastic optimization, the learning based solution is unable to satisfy the QoS requirement with probability $1$. In the sequel, we evaluate the availability, which is the probability that the QoS requirement in \eqref{con:QLnr} can be satisfied. Since both the minimal bandwidth and the convergence rate are insensitive to the reliability requirement, we have a chance to improve the availability with minor sacrifice of bandwidth and higher training complexity. In particular, we reduce the overall reliability requirement during training, denoted as $\varepsilon_\mathrm{D}$, from the required overall reliability, i.e., $\varepsilon_\mathrm{D} \!<\!\varepsilon_\mathrm{max}$. With such a conservative design, even if the overall packet loss probability achieved by the learnt solution violates $\varepsilon_\mathrm{D}$, it is less likely to violate the original reliability requirement $\varepsilon_\mathrm{max}$.
To evaluate the bandwidth loss, we define the $1$st percentile bandwidth loss as $\tilde{W} \!\triangleq\! \inf {\{x | \Pr{\{\hat{W} \!-\! W^*\}} \!\leq\! 1\%\}}$, which means that the bandwidth loss is lower than $\tilde{W}$ in $99\%$ cases.

\begin{table}[htbp]
	\vspace{-0.2cm}
	\small
	\renewcommand{\arraystretch}{1.3}
	\caption{Availability and Bandwidth Loss in Ensuring $\varepsilon_\mathrm{max} \!=\! 10^{-5}$. }	\label{tab:Avail}
	\begin{center} \vspace{-0.4cm}
    \begin{tabular}{|l|c|c|c|}
    \hline
    $\varepsilon_\mathrm{D}$ & $10^{-5}$ & $10^{-6}$ & $10^{-7}$ \\ \hline
    Availability & $46.7\%$ & $98.9\%$ & $99.96\%$ \\ \hline
    $\tilde{W}$ & $1.3\%$ & $3.3\%$ & $5.7\%$ \\ \hline
    \end{tabular}
    \end{center}	\vspace{-0.5cm}
\end{table}

Table \ref{tab:Avail} shows the availability and the $1$st percentile bandwidth loss for different values of $\varepsilon_\mathrm{D}$. The results are obtained with $1\,000$ times of training, each with $10\,000$ iterations. The bandwidth loss and the availability of each training result are evaluated over $1\,000$ randomly generated large-scale channel gains. We can see that setting $\varepsilon_\mathrm{D} \!=\! \varepsilon_\mathrm{max}$ achieves low availability, which actually is always around $50\%$ even with more iterations. This is because the optimal solution of the problem in \eqref{prob:RA} is obtained when the equality in \eqref{con:QLnr} holds, which cannot be strictly achieved with stochastic optimization. Nonetheless,
by reducing $\varepsilon_\mathrm{D}$, the availability can be remarkably improved with a minor increase of the minimal bandwidth.
Besides, the complexity of training for achieving high availability is not high. Less than $10$~s is required for $10\,000$ iterations by a computer with {Intel\textregistered}\ {Core\texttrademark}\ {i9-9920X} CPU without using the acceleration from GPU.

\section{Conclusion}
In this paper, we proposed an approach to learn the relation between the solution of constrained-optimization problem and system parameters with unsupervised learning. The problem can either be variable optimization or functional optimization. We illustrated the approach by considering bandwidth minimization problem under the QoS constraint of URLLC. Simulation results showed that the convergence speed of the proposed unsupervised learning is insensitive to the required reliability, and the solution learnt by DNN can achieve high availability with a minor loss of the bandwidth efficiency.

\appendices
\section{Proof of the Equivalence between Problems \eqref{prob:VarOpt} and \eqref{prob:VarOptb}}
\label{App:Upgrade}
\renewcommand{\theequation}{A.\arabic{equation}}
\setcounter{equation}{0}
\begin{proof}
Since $\mathcal{F}[\cdot]$ has positive variation,  if $f\left(\bm{x}_1 (\bm{\theta});\bm{\theta}\right) \!\leq\! f\left(\bm{x}_2 (\bm{\theta});\bm{\theta}\right)$ for any values of $\bm{\theta}$, then $\mathcal{F} [f\left(\bm{x}_1 (\bm{\theta});\bm{\theta}\right)] \!\leq\! \mathcal{F} [f\left(\bm{x}_2 (\bm{\theta});\bm{\theta}\right)]$.
Since the problems in \eqref{prob:VarOpt} and \eqref{prob:VarOptb} have the same constraints, they have the same feasible area. Therefore, $\bm{x}^* (\bm{\theta})$ is feasible for the problem in \eqref{prob:VarOptb}.
Then, in order to prove the equivalence of the two problems we only need to prove that for each feasible solution of the problem in \eqref{prob:VarOptb}, $\bm{x}_0 (\bm{\theta})$, we have $\mathcal{F} [f\left(\bm{x}^* (\bm{\theta});\bm{\theta}\right)] \!\leq\! \mathcal{F} [f\left(\bm{x}_0 (\bm{\theta});\bm{\theta}\right)]$.

Since $\bm{x}_0 (\bm{\theta})$ is also feasible for the problem in \eqref{prob:VarOpt} and $\bm{x}^*(\bm{\theta})$ is optimal for the problem in \eqref{prob:VarOpt}, we have,
\begin{align}   \label{opt:VarOpt}
  f\left(\bm{x}^* (\bm{\theta});\bm{\theta}\right) \leq f\left(\bm{x}_0 (\bm{\theta});\bm{\theta}\right), \  \forall \bm{\theta} \in \mathcal{A}.
\end{align}
Then, considering that $\mathcal{F}[\cdot]$ is a monotonically increasing function, we have $\mathcal{F} [f\left(\bm{x}^* (\bm{\theta});\bm{\theta}\right)] \!\leq\! \mathcal{F} [f\left(\bm{x}_0 (\bm{\theta});\bm{\theta}\right)]$. This completes the proof.
\end{proof}

\bibliographystyle{IEEEtran}
\bibliography{ref}

\end{document}